\documentclass[10pt,twocolumn,letterpaper]{article}

\usepackage[pagenumbers]{cvpr} % To force page numbers, e.g. for an arXiv version

\usepackage{graphicx}
\usepackage{amsmath}
\usepackage{amssymb}
\usepackage{booktabs}
\usepackage{float}
\usepackage{multirow}
\usepackage{array}
\usepackage{caption}

\usepackage{times}
\usepackage{epsfig}
\usepackage{algorithm,algpseudocode}

\usepackage{color}
\usepackage{booktabs}
\usepackage{bbding}
\usepackage{multirow}

\usepackage{pifont}% http://ctan.org/pkg/pifont
\usepackage{subcaption}
\usepackage[pagebackref,breaklinks,colorlinks]{hyperref}

\usepackage[capitalize]{cleveref}
\crefname{section}{Sec.}{Secs.}
\Crefname{section}{Section}{Sections}
\Crefname{table}{Table}{Tables}
\crefname{table}{Tab.}{Tabs.}

\def\cvprPaperID{6783} % *** Enter the CVPR Paper ID here
\def\confName{CVPR}
\def\confYear{2023}

\definecolor{mygreen}{RGB}{0,176,80}

\makeatletter
\newcommand{\thickhline}{%
    \noalign {\ifnum 0=`}\fi \hrule height 1pt
    \futurelet \reserved@a \@xhline
}
\newcolumntype{"}{@{\hskip\tabcolsep\vrule width 1pt\hskip\tabcolsep}}
\makeatother

\begin{document}

\title{Spatial-then-Temporal Self-Supervised Learning for Video Correspondence}

% \and
\author{Rui Li ~~~~~~~~~~~~~~~~~~~~~~~~~ Dong Liu  \\
         University of Science and Technology of China, Hefei, China\\
{\tt\small liruid@mail.ustc.edu.cn, \tt\small dongeliu@ustc.edu.cn}
}

{
	\maketitle
	\thispagestyle{empty}
}

%%%%%%%%% ABSTRACT
\begin{abstract}
  \footnotetext[1]{This work was supported by the Natural Science Foundation of China under Grants 62022075 and 62036005, and by the Fundamental Research Funds for the Central Universities under Contracts WK3490000005 and WK3490000006. \emph{(Corresponding author: Dong Liu.)}}
  In low-level video analyses, effective representations are important to derive the correspondences between video frames. These representations have been learned in a self-supervised fashion from unlabeled images or videos, using carefully designed pretext tasks in some recent studies. However, the previous work concentrates on either spatial-discriminative features or temporal-repetitive features, with little attention to the synergy between spatial and temporal cues. To address this issue, we propose a spatial-then-temporal self-supervised learning method. Specifically, we firstly extract spatial features from unlabeled images via contrastive learning, and secondly enhance the features by exploiting the temporal cues in unlabeled videos via reconstructive learning. In the second step, we design a global correlation distillation loss to ensure the learning not to forget the spatial cues, and a local correlation distillation loss to combat the temporal discontinuity that harms the reconstruction. The proposed method outperforms the state-of-the-art self-supervised methods, as established by the experimental results on a series of correspondence-based video analysis tasks. Also, we performed ablation studies to verify the effectiveness of the two-step design as well as the distillation losses. Our code and models are available at \textcolor{magenta}{https://github.com/qianduoduolr/Spa-then-Temp}.
\end{abstract}

%%%%%%%%% BODY TEXT
\section{Introduction}
Learning representations for video correspondence is a fundamental problem in computer vision, which is closely related to different downstream tasks, including optical flow estimation~\cite{dosovitskiy2015flownet, horn1981determining}, video object segmentation~\cite{caelles2017one, oh2019video}, keypoint tracking~\cite{xiu2018pose}, etc. However, supervising such a representation requires a large number of dense annotations, which is unaffordable. Thus, most approaches acquire information from simulations~\cite{dosovitskiy2015flownet, mayer2016large} or limited annotations~\cite{pont20172017, xu2018youtube}, which result in poor generalization in different downstream tasks. Recently, self-supervised feature learning is gaining significant momentum. Several pretext tasks~\cite{he2020momentum, xie2021detco, jabri2020space,lai2020mast, li2019joint,wang2019learning} are designed, mostly concentrating on either spatial feature learning or temporal feature learning for space-time visual correspondence.

\begin{figure}[!t]
  \centering
  \includegraphics[width=1.0\linewidth]{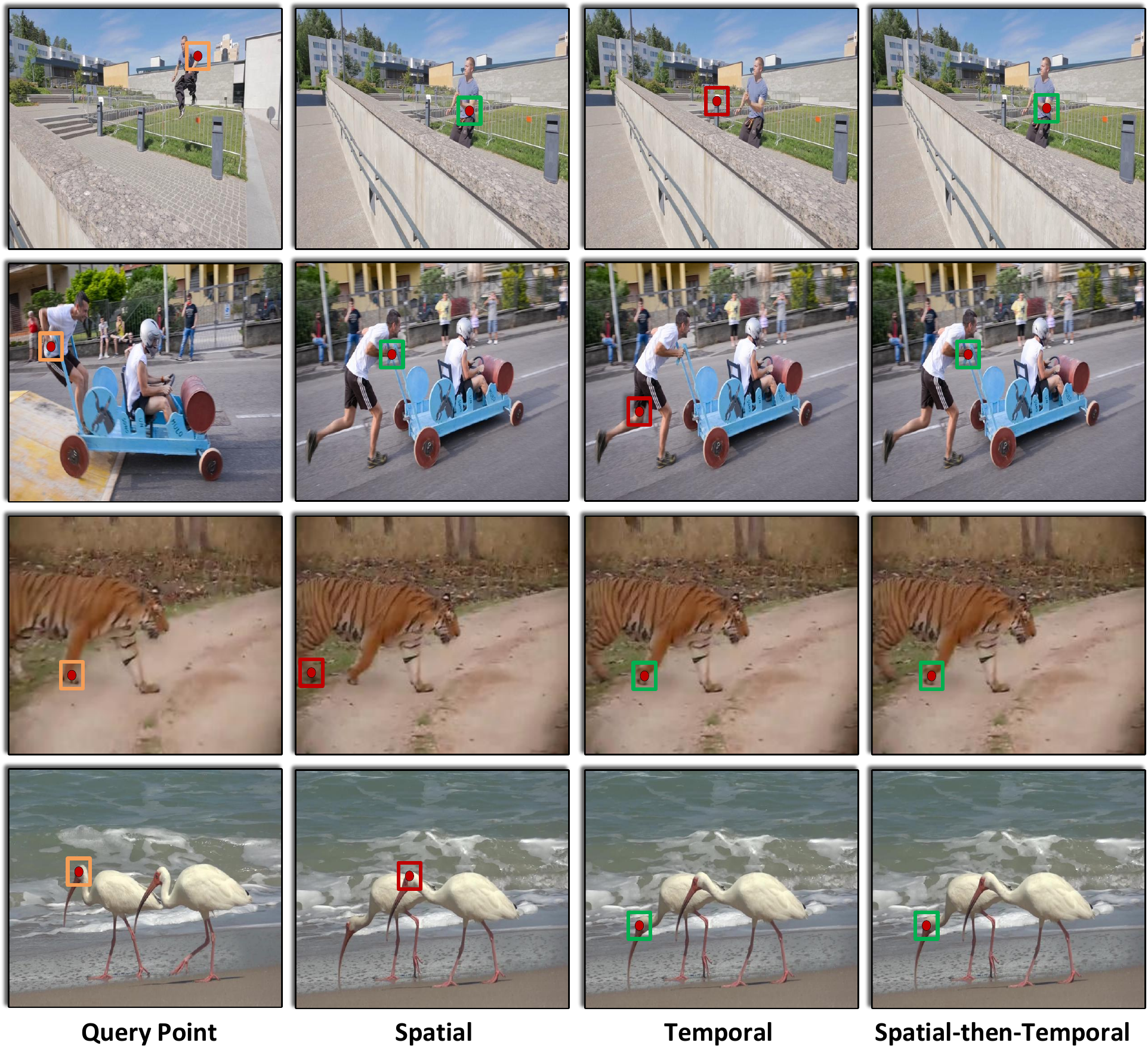}
  \vspace{-7mm}
  \captionof{figure}{\textbf{Matching results given the query point.} While temporal-repetitive features fail to handle video correspondence with dramatic appearance changes and deformations~(the first two rows), spatial-discriminative features are incompetent to recognize the temporal repetition and are misled by distractors with similar appearance~(the last two rows). The green/red bounding boxes indicate correct/wrong matching.~(\textbf{Zoom in for best view})}
  \vspace{-7mm}
  \label{fig:teaser}
\end{figure}%

With the objective of learning the representations that are invariant to the appearance changes, \textbf{spatial feature learning} provides video correspondence with discriminative and robust appearance cues, especially when facing severe temporal discontinuity, i.e., occlusions, appearance changes, and deformations. Most recently, as mentioned in~\cite{wang2021different}, the contrastive models~\cite{he2020momentum, xie2021detco,chen2020simple} pre-trained on image data show competitive performance against dedicated methods for video correspondence. Thus, it is shown to be a better way for learning spatial representations in terms of quality and data efficiency, compared with the methods~\cite {jabri2020space, li2019joint, wang2020contrastive, xu2021rethinking, zhao2021modelling,son2022contrastive} using large-scale video datasets for training. Though getting promising results, as shown in Figure~\ref{fig:teaser}, the learned spatial representations are misled by the distractors with similar appearance, which indicates the poor ability to recognize the temporal pattern. 

In another line, \textbf{temporal feature learning} focuses on learning the temporal-repetitive features that occur consistently over time. With the temporal consistency assumption~\cite{black1993framework}, the pixel repetition across video motivates recent studies to exploit the temporal cues via a reconstruction task~\cite{meister2018unflow, liu2019ddflow,yu2016back,lai2020mast}, where the query pixel in the target frame can be reconstructed by leveraging the information of adjacent reference frames within a local range. Then a reconstruction loss is applied to minimize the photometric error between the raw frame and its reconstruction. Nevertheless, temporal-repetitive features highly depend on the consistency of the pixels across the video. It thus can be easily influenced by the temporal discontinuity caused by the dramatic appearance changes, occlusions, and deformations~(see Figure~\ref{fig:teaser}).

In light of the above observation, we believe the video correspondence relies on both spatial-discriminative and temporal-repetitive features. Thus, we propose a novel spatial-then-temporal pretext task to achieve synergy between spatial and temporal cues. Specifically, we firstly learn spatial features from unlabeled images via contrastive learning and secondly improve the features by exploiting the temporal repetition in unlabeled videos with frame reconstruction. While such an implementation brings together the advantages of spatial and temporal feature learning, there are still some problems. First of all, the previous studies~\cite{jabri2020space, li2019joint, wang2020contrastive, xu2021rethinking, hu2022semantic} propose to learn coarse-grained features for video correspondence. With the objective of reconstructive learning, the video frames need to be down-sampled to align with the coarse-grained features~\cite{lai2020mast}, resulting in severe temporal discontinuity for the pixels to be reconstructed. Thus, the frame reconstruction loss becomes invalid. Second, in the context of sequential training, directly training with only new data and objective functions will degrade the discriminative features learned before.

To tackle the first problem, we firstly exploit temporal cues by frame reconstruction at different pyramid levels of the encoder.  We observe that temporal repetition benefits from relatively smaller down-sampling rate. Hence, the model will learn better temporal-persistent features at the fine-grained pyramid level.  To distillate the knowledge from it, we design a local correlation distillation loss that supports explicit learning of the final correlation map in the region with high uncertainty, which is achieved by taking the more fine-grained local correlation map as pseudo labels. This leads to better temporal representations on the coarse feature map. At the same time, we regard the model pre-trained in the first step as the teacher. Then a global correlation distillation loss is proposed to retain the spatial cues. Eventually, we can obtain better temporal representations without losing the discriminative appearance cues acquired in the first step.

To sum up, our main contributions include: (i)~We propose a spatial-then-temporal pretext task for self-supervised video correspondence, which achieves synergy between spatially discriminative and temporally repetitive features. (ii)~We propose the local correlation distillation loss to facilitate the learning of temporal features in the second step while retaining the appearance sensitivity learned in the first step by the proposed global correlation distillation loss. (iii)~We verify our approach in a series of correspondence-related tasks. Our approach consistently outperforms previous state-of-the-art self-supervised methods and is even comparable with task-specific fully-supervised algorithms.

\section{Related Work}

\begin{figure*}[!tb]
  \centering
  {\includegraphics[width=0.91\textwidth]{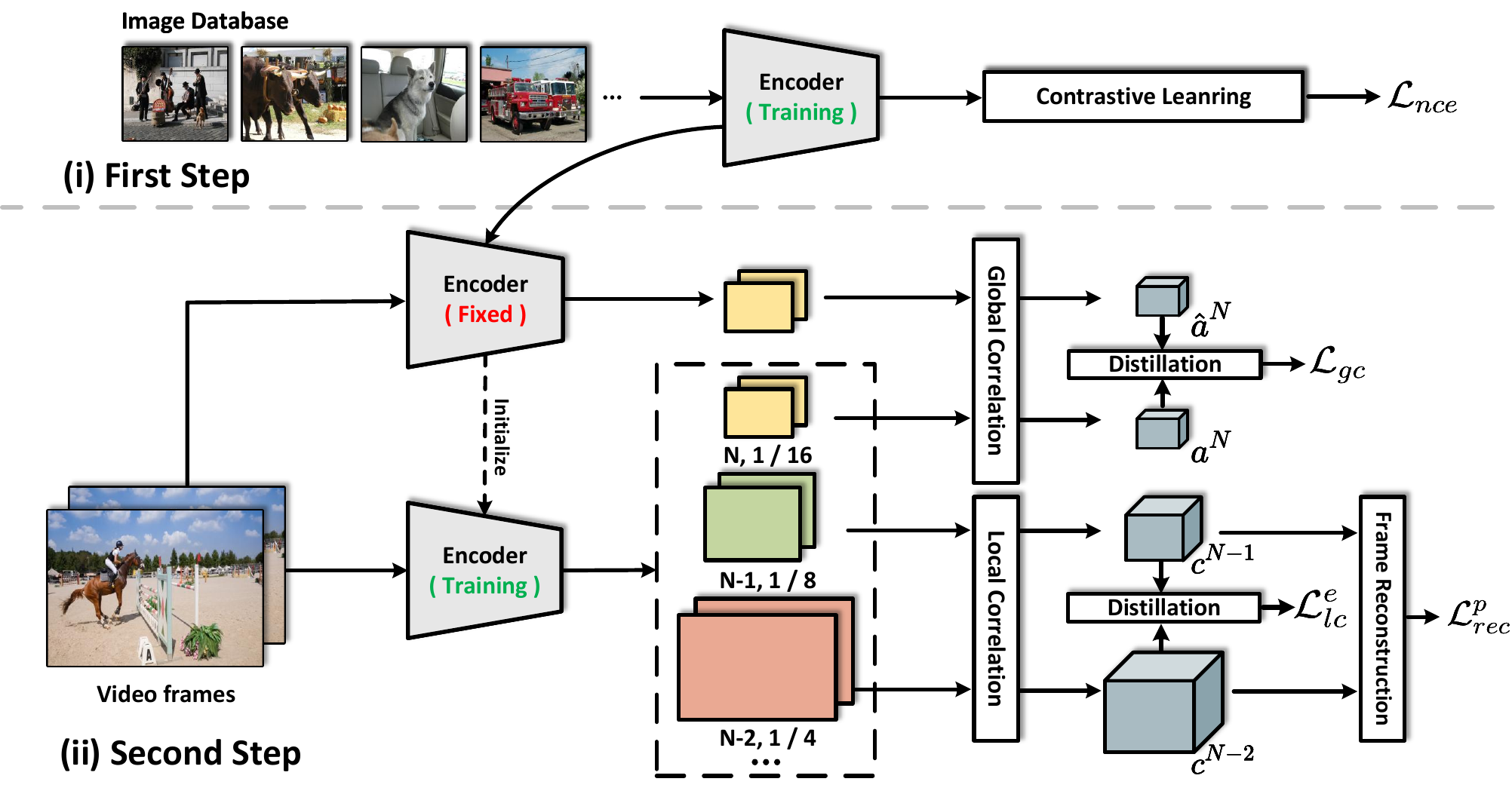}}
  \vspace{-3mm}
  \caption{\small \textbf{Overview of the two-step design in our spatial-then-temporal pretext task. In the second step, we fix the encoder trained in the first step as the teacher.}  Following prior studies~\cite{jabri2020space, wang2020contrastive, xu2021rethinking}, we focus on learning dense features at the level $N-1$~(the level indicates the layer of the ResNet architecture). Based on discriminative appearance features already learned in the first step, we exploit temporal cues at each pyramid level of the encoder by reconstructive learning. Each query pixel in the target frame can be reconstructed by a weighted sum of pixels within a local window in the reference frame. Then, we distillate the knowledge from a more fine-grained pyramid level~(i.e. $N-2$) by aligning the local correlation maps in the region with high uncertainty, which leads to better temporal features at level $N-1$. Meanwhile, we compute the global correlation to take account into the large motion between frames, and a novel global correlation distillation loss is devised between global correlation maps to retain spatially discriminant features learned in the first step.}
  \label{fig:framework}
  \vspace{-6mm}
\end{figure*}

\textbf{Self-supervised representation learning for video correspondence.} 
Recent approaches focus on learning the dense representations from unlabeled videos in a self-supervised manner for video correspondence, which has proceeded along two different dimensions: reconstruction-based methods~\cite{lai2019self, lai2020mast, li2019joint, vondrick2018tracking, wang2020contrastive} and cycle-consistency-based methods \cite{jabri2020space, wang2019learning, zhao2021modelling}. In the first type, a query point is reconstructed from adjacent frames while the latter performs forward-backward tracking with the objective of minimizing the cycle inconsistency. Though getting promising results, the learned features still lack the capability of spatial discriminating. Meanwhile, VFS~\cite{xu2021rethinking} try to learn the spatial and temporal representation through a frame-wise contrastive loss while the methods in~\cite{araslanov2021dense, wang2020contrastive} try to realize it by exploiting the inter-video constraint, which may result in sub-optimal performance. The concurrent work SFC~\cite{hu2022semantic} proposes a two-stream network that learns the semantic and fine-grained features with two separate models. Compared to SFC~\cite{hu2022semantic} which does not consider the temporal pattern, our method is able to learn both spatial and temporal features with only a single model.

\textbf{Self-supervised optical flow for video correspondence.} 
Apart from learning powerful representations for video correspondence, another line of studies~\cite{dosovitskiy2015flownet, teed2020raft} approaches the problem by directly regressing the optical flow produced by synthetic datasets, thus suffering from severe domain shifts. To address the problem, the study in~\cite{yu2016back} first introduces a method for learning optical flow without using any annotations by leveraging brightness constancy with photometric reconstruction loss, and Unflow~\cite{meister2018unflow} further improves it with occlusions reasoning. Though making great progress, it still does not generalize well in real scenarios. In~\cite{jonschkowski2020matters, liu2020learning, liu2019ddflow}, they improve performance by utilizing the optical flow predictions from the teacher model to guide the learning of the student model in the region with occlusions. In this paper, we propose a novel self-distillation loss to combat temporal discontinuity in the context of self-supervised feature learning.

\textbf{Self-supervised spatial representation learning.} Self-supervised spatial feature learning recently gets the promising result with contrastive learning. In an early work~\cite{wu2018unsupervised}, contrastive learning is formulated as an instance discrimination task, which requires the model to return low values for similar pairs and high values for dissimilar pairs. The performance is further improved by creating a dynamic memory-bank~\cite{he2020momentum}, introducing online clustering~\cite{caron2020unsupervised}, and avoiding the use of negative pairs~\cite{chen2021exploring, grill2020bootstrap}. Furthermore, the methods in \cite{wang2021dense, xie2021detco, yang2021instance} propose various pretext tasks to adapt the contrastive learning to dense prediction tasks. Even though showing superior performance for video correspondence~\cite{wang2021different}, the contrastive model still struggles to figure out the temporal pattern between video frames. 

\vspace{-2mm}

\section{Method}
In this section, we firstly introduce spatial feature learning in the context of contrastive learning. Then, the temporal feature learning is improved with our pyramid frame reconstruction and local correlation distillation. In the end, we present our spatial-then-temporal pretext task along with the global correlation distillation used for retaining the learned spatial cues in the first step.

\vspace{-1mm}

\subsection{Spatial Feature Learning}
\label{spatial_feature_learning}
Spatial feature learning aims to learn the discriminative features of different objects, which provides video correspondence with robust appearance cues, especially when facing temporal discontinuity. Most recently, spatial feature learning is gaining significant momentum due to the advancements in contrastive learning. We begin by briefly reviewing instance discrimination objective in contrastive learning. Specifically, the query vector $\boldsymbol{q}\in\mathbb{R}^d$ and a set of key vectors $\mathcal{K}=\left\{\boldsymbol{k}^{+}, \boldsymbol{k}_{1}^{-}, \boldsymbol{k}_{2}^{-}, \ldots, \boldsymbol{k}_{K}^{-}\right\}$ which consists of one positive key $\boldsymbol{k}^{+}\in\mathbb{R}^d$ and $K$ negative keys $\mathcal{K^{-}}=\left\{\boldsymbol{k}_{j}^{-}\right\}$, are encoded by the encoder $\phi$ plus an MLP head, and are further processed by $l_2$-normalization. The query and its positive key are generated from the same image with two different augmentations, while the negative keys refer to other instances. The objective of instance discrimination is to maximize the similarity between the query $\boldsymbol{q}$ and the positive key $\boldsymbol{k}^{+}$ while the remaining query distinct to all negative keys $\mathcal{K^{-}}$. Thus, a contrastive loss is presented in InfoNCE~\cite{oord2018representation}  with a softmax formulation:
  \begin{equation}\label{eq:nce}
    \small
    \begin{aligned}
      \mathcal{L}_{\text{nce}}=-\log \frac{\exp \left(\boldsymbol{q}^{T} \cdot \boldsymbol{k}^{+} / \tau_c\right)}{\exp \left(\boldsymbol{q}^{T} \cdot \boldsymbol{k}^{+} / \tau_c\right)+\sum_{i=1}^{K} \exp \left(\boldsymbol{q}^{T} \cdot\boldsymbol{k}_{i}^{-} / \tau_c\right)}
    \end{aligned}~~,
  \end{equation}
where the similarity is measured via dot product, and $\tau_c$ is the temperature hyper-parameter. MoCo~\cite{he2020momentum} builds a dynamic memory bank to maintain a large number of negative samples with a moving-averaged encoder. DetCo~\cite{xie2021detco} further improves the contrastive loss $\mathcal{L}_{\text{nce}}$ by introducing global and local contrastive learning to enhance local representations. In this paper, we adopt DetCo  to learn appearance features for most of our experiments.

\begin{figure}[!tb]
  \centering
  {\includegraphics[width=0.43\textwidth]{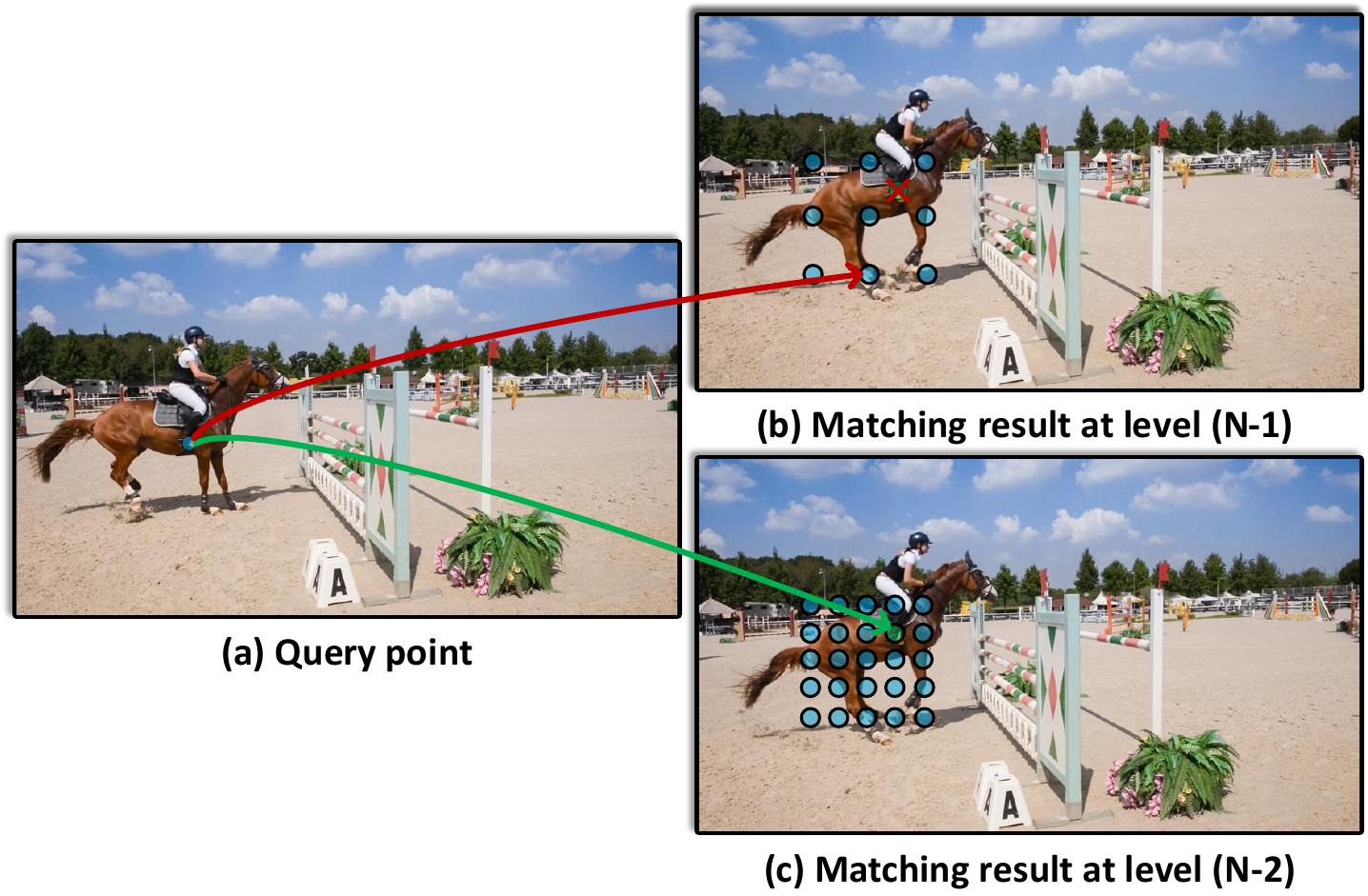}}
  \vspace{-2mm}
  \caption{\small \textbf{Visualization of the local correlation of the query point at different pyramid levels.} The red arrow indicates the wrong match given the query point. In (b), the frame reconstruction becomes invalid due to the absence of the query point in the reference frame. (\textbf{Zoom in for best view})}
  \label{fig:temp}
  \vspace{-5mm}
\end{figure}

\subsection{Temporal Feature Learning}
\label{temporal_feature_learning}
Temporal feature learning aims to learn temporally repetitive features between video frames, which is closely related to the recent studies of motion estimation~\cite{teed2020raft,dosovitskiy2015flownet,liu2019ddflow}. Based on the temporal consistency assumption~\cite{black1993framework}, the pixel repetition encourages the studies in~\cite{wang2019learning, meister2018unflow, lai2020mast,liu2019ddflow} to learn the video correspondence by reconstructive learning, where each query pixel in the target frame can be reconstructed by leveraging the information of adjacent reference frames with a limited range. More specifically, the target and reference frame $I_{t}, I_r \in \mathbb{R}^{H \times W \times 3}$ are projected into a pyramid pixel embedding space by the encoder $\phi$. We denote these embeddings as $F^l_t,F^l_{r} \in \mathbb{R}^{h^l w^l \times d^l}$, where $l \in \{0, 1, \ldots, N \}$ is the index of each pyramid level and the bigger number represents the coarser pyramid level. The $s^l$ denotes the stride up to each pyramid level and $h^l = H/s^l, w^l = W/s^l$. From this point of view, each feature point is aligned with the center pixel of strided convolution layers~\cite{lai2020mast} and we can sample it to get the aligned target and reference frame $\hat{I}^l_{t}, \hat{I}^l_r \in \mathbb{R}^{h^lw^l \times 3}$. For each query pixel $i$ in $\hat{I}^l_{t}$ at the pyramid level $l$,  we can calculate the local correlation~\cite{lai2019self} $c^l(i,j)$ w.r.t. each key pixel $j$ in the reference frame within a local window centered at $i$, considering the nature of temporal coherence in the video:
\begin{equation}\label{eq:local_correlation}
  \footnotesize
  \begin{aligned}
    c^l(i,j)=\frac{\exp \left(F^l_t(i)  \cdot F^l_r(j) / \tau\right)}{\sum_{n} \exp \left(F^l_t(i) \cdot F^l_r(n) / \tau\right)}, i \in \{1,.,h^lw^l\}, j,n \in \mathcal{N}(i)
  \end{aligned}~~,
\end{equation}
where $\mathcal{N}(i)$ is the index set in the reference frame with a limited range of $R^l$ for the pixel $i$. Then the query pixel $i$ in the target frame can be reconstructed by a weighted sum of pixels in $\mathcal{N}(i)$, according to the local correlation map $c^l \in \mathbb{R}^{h^lw^l \times (2R^l+1)^2}$:
\begin{equation}\label{eq:reconstruction}
  \small
  \begin{aligned}
    \overline{I}^l_{t}(i)=\sum_{j \in \mathcal{N}(i)} c^l(i,j) \hat{I}^l_{r}(j)
  \end{aligned}~~.
\end{equation}
 Then the reconstruction loss $\mathcal{L}_{\mathrm{rec}}$ is defined as $L_1$ distance between $\hat{I}^l_{t}$ and $\overline{I}^l_{t}$ at the pyramid level $l$:
\begin{equation}\label{eq:reconstruction loss}
  \small
  \begin{aligned}
    \mathcal{L}_{\mathrm{rec}}=\left\|\hat{I}^l_{t} - \overline{I}^l_{t}\right\|_{1}
  \end{aligned}~~.
\end{equation}
since the previous methods focus on learning dense representations at level $N-1$, we set $l$ to $N-1$ by default, which is regarded as our baseline.

\vspace{1mm}
\textbf{Pyramid frame reconstruction.}  We can further exploit more fine-grained temporal cues of the raw frames. As observed in Figure \ref{fig:framework}, we obtain a pair of pyramid features $\{F^l_t\}^{N-1}_{l=1}$,$\{F^l_{r}\}^{N-1}_{l=1}$, which is utilized to calculate the pyramid local correlation maps $\{c^l\}^{N-1}_{l=1}$ with Eq.~(\ref{eq:local_correlation}). Then, we can reconstruct the target frame at each pyramid level using Eq.~(\ref{eq:reconstruction}). We devise the pyramid reconstruction loss as:
\begin{equation}\label{eq:pyramid reconstruction loss}
  \small
  \begin{aligned}
    \mathcal{L}^{p}_{\mathrm{rec}}=\sum_l\left\|\hat{I}^l_{t} - \overline{I}^l_{t}\right\|_{1}
  \end{aligned}~~.
\end{equation}

\textbf{Local correlation distillation.} For video correspondence, previous methods learn the coarse-grained features with large down-sampling rate, which would harm the pixel repetition between video frames. As observed in Figure~\ref{fig:temp}, the query pixel in the target frame is absent in level $N-1$, resulting in an obvious shortcut that the query pixel tries to match any other pixel with a similar color to reduce the frame reconstruction loss~(red arrow). However, the missing pixel can be found at level $N-2$ with higher resolution~(see Figure~\ref{fig:temp}~(c)). Hence, the temporal feature learning performed at this pyramid level would contribute to better temporal features. Here, we design a novel local correlation distillation loss to distillate the knowledge from more fine-grained pyramid levels. Specifically, we firstly leverage correlation down-sampling~\cite{teed2020raft} on $c^{N-2}$ to get pseudo labels $\hat{c}^{N-1}$ with the same size as $c^{N-1}$.  Then we propose a local correlation distillation loss $\mathcal{L}_{lc}$ to minimize the mean squared error between them. By applying the novel self-distillation, we can eventually learn better temporal-related features at relatively coarse pyramid level with low computational costs of inference.

\textbf{Entropy-based selection.} The correlation of each query w.r.t. reference frame indicates more uncertainty when having smooth distribution, which has a higher probability of suffering the temporal discontinuity. Thus, it should be paid more attention to when applying local correlation distillation. We calculate the entropy for each query $i$:
\begin{equation}\label{eq:entropy}
  \small
  \begin{aligned}
    \mathcal{H}(i)=\sum_j -\log c^{N-1}(i,j),  i \in \{1,.,h^{N-1}w^{N-1}\}, j \in \mathcal{N}(i)
  \end{aligned}~~,
\end{equation}
then we obtain a mask $m \in \{0,1\}^{h^{N-1}w^{N-1}}$ to filter out the region with lower entropy by setting a threshold $T$. The local correlation distillation loss with entropy selection is defined as:
\begin{equation}\label{eq:local_correlation_dist}
  \small
  \begin{aligned}
    \mathcal{L}^e_{\mathrm{lc}}=\sum_im(i) \cdot \left\|c^{N-1}(i)-\hat{c}^{N-1}(i)\right\|^2_{2}
  \end{aligned}~~.
\end{equation}
Eventually, our training loss of temporal feature learning is defined as $\mathcal{L}_\mathrm{t}$ =  $\mathcal{L}^p_{\mathrm{rec}} + \alpha  \mathcal{L}^e_{\mathrm{lc}}$. 

\vspace{1mm}
\begin{figure}[!tb]
  \centering
  {\includegraphics[width=0.47\textwidth]{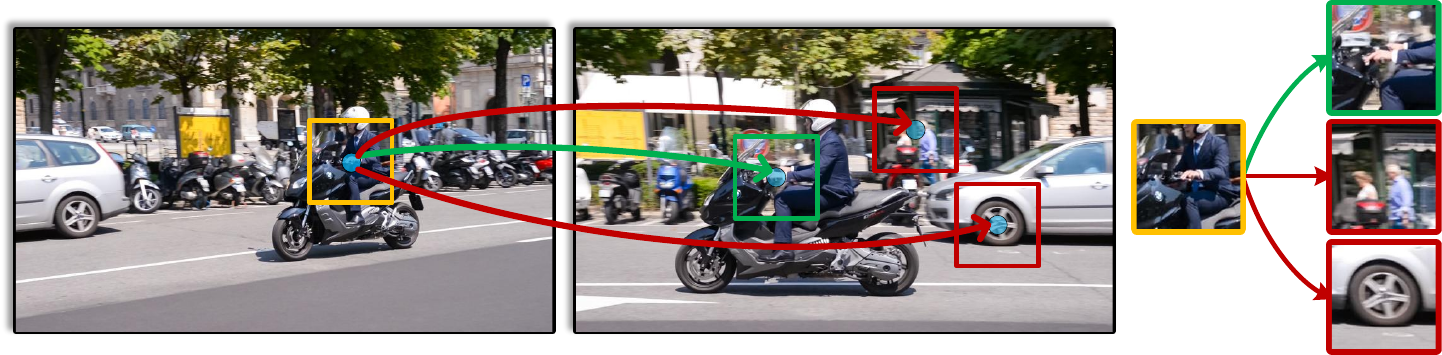}}
  \vspace{-2mm}
  \caption{\small \textbf{Visualization of the global correlation of the query point for large motion.} The red arrow indicates the wrong match given the query point, and the box indicates the patch centered at each point. The model trained in the first step can handle this case with learned discriminative appearance features.  }
  \label{fig:spa}
  \vspace{-6mm}
\end{figure}

\subsection{Spatial-then-Temporal Feature Learning}\label{configurations}
In this section, we will formally introduce our spatial-then-temporal self-supervised learning for video correspondence. Besides, we further devise some variants of training configurations for learning the spatiotemporal features for comparisons.
\vspace{1mm}

(i) \textbf{Spatial-then-Temporal}: 
% In such an implementation,    Apart from the temporal discontinuity coming from the down-sampling, as shown in Figure~\ref{fig:spa}, the changes of the appearance and illumination also prevent the model from learning appropriate correspondence, i.e., the green arrow, since the temporal consistency assumption does not hold and the query point may match to other distractors at any other position in the reference frame with similar color (e.g. the red arrow).  This issue has already been addressed in the first stage of training with robust appearance features. 
We first train the encoder $\phi$ with unlabeled image data via contrastive learning. Next, we concentrate on learning temporal-related features with $\mathcal{L}_t$ on the video dataset. However, the model will not be able to capture spatial cues when training it with new object functions and data.  Consequently, it may fail to deal with the dramatic appearance changes and deformations with large motion like before~(see Figure~\ref{fig:spa}). Thus, we introduce a global correlation distillation loss to tackle the problem.  Precisely, we first fix the feature encoder $\phi$ trained in the first step as the teacher denoted as $\phi_t$. For each query point $F^N_t(i)$ at pyramid level $N$, we compute the global correlation $a^N(i, j)$ using a softmax over similarities w.r.t. all keys in the reference frame, i.e., 
\begin{equation}\label{eq:global_correlation}
  \footnotesize
  \begin{aligned}
    a^N(i, j)=\frac{\exp \left(F^N_t(i)  \cdot F^N_r(j) / \tau\right)}{\sum_{n} \exp \left(F^N_t(i) \cdot F^N_r(n) / \tau\right)}, i,j,n \in \{1,\ldots,h^Nw^N\}
  \end{aligned}~~,
\end{equation}
     we generate the pseudo labels $\hat{a}^N$  using Eq.~(\ref{eq:global_correlation}) with $\phi_t$. The global correlation distillation loss $\mathcal{L}_{\mathrm{gc}}$ is defined to minimize the mean squared error between $a^N$ and $\hat{a}^N$:
\begin{equation}\label{eq:global_correlation_loss}
  \small
  \begin{aligned}
    % \mathcal{L}_{gc}  =\sum_{i}\sum_{j} \left\|a_{i,j}-a^t_{i,j}\right\|_{2}^{2}
    \mathcal{L}_{\mathrm{gc}}  = \left\|a^N-\hat{a}^N\right\|_{2}^{2}
  \end{aligned}~~,
\end{equation}
by aligning the $a^N$ with $\hat{a}^N$, the student tries to retain the appearance features learned in the first step. It is worth noting that compared with the local correlation that only considers the video correspondence in a local range, the global correlation compares the query with all instances in the reference frame. Thus, it also provides more valuable training samples for distilling the spatially discriminant features with $\mathcal{L}_{gc}$.  The final loss of training on video data is $ \mathcal{L} = \mathcal{L}_{\mathrm{t}}  + \beta  \mathcal{L}_{\mathrm{gc}}$

\vspace{1mm}

(ii) \textbf{Spatiotemporal}: We can also learn the spatial and temporal features jointly with both $\mathcal{L}_{nce}$ and $\mathcal{L}_{t}$. The training loss is summarized as  $\mathcal{L} = \mathcal{L}_{\mathrm{nce}}  + \gamma \mathcal{L}_{\mathrm{t}}$.

\vspace{1mm}

(iii) \textbf{Temporal-then-Spatial}: Then, we try to reverse the training order of (i). However, without learning the robust appearance features first, temporal feature learning may be misled by the distractors when facing severe temporal discontinuity. 

% However, such a design brings about two drawbacks. First, the temporal information can not be accessed at the second stage since we perform spatial feature learning on unlabeled image data, which makes it difficult to design a loss like $\mathcal{L}_{gc}$. Second, without learning the robust appearance features, temporal feature learning may be misled by the distractors when facing severe temporal discontinuity.

The comparisons between different training configurations will be discussed in the experiments, and the spatial-then-temporal is used as the default training configuration.

\section{Experiments}\label{exp}
We verify the merit of our method in a series of correspondence-related tasks, including semi-supervised video object segmentation, human part propagation, and pose keypoint tracking. This section will firstly introduce our experiment settings, including implementation and evaluation details. Then detailed ablation studies are performed to explain how each component of our method works. We finally report the performance comparison with state-of-the-art methods to further verify the effectiveness of our method.

\subsection{Implementation Details}
\textbf{Backbone.} We exploit the encoder $\phi$ with both ResNet-18 and ResNet-50~\cite{he2016deep} for self-supervised training. When training at the first step, the encoder plus an MLP head project each image to a vector. Then, for temporal feature learning, we reduce the stride up to layer res$_4$ from 16 to 8~(in Figure~\ref{fig:framework}) to increase the spatial resolution of feature maps by a factor of 2. We further reduce the stride to 4 to compare with prior studies~\cite{li2022locality,miao2022self,lai2020mast} which perform self-supervised training on high-resolution feature maps.

\textbf{Training details.}
We first train our model using contrastive loss for 200 epochs on ImageNet~\cite{deng2009imagenet} following most augmentations and hyper-parameters settings of~\cite{he2020momentum,xie2021detco}. Then we perform temporal feature learning on the training set of YouTube-VOS~\cite{xu2018youtube} which consists of 3.5k videos. In this stage, the video frame is resized into 256$\times$256, and channel-wise dropout with probability of 0.5 in Lab color space~\cite{lai2019self, lai2020mast} is adopted as the information bottleneck. We train the encoder with the stride of 8 for 90k/45k iterations with a mini-batch of 128/64 for ResNet-18/ResNet-50,  using Adam as our optimizer. The initial learning rate is set to 1e-4 with a cosine (half-period) learning rate schedule. The range~$R$ of computing the local correlation map is set to 12/6 on layer res$_3$/res$_4$. The experiments for temporal feature learning are performed with 4 GTX-3090 GPUs. We include more training details in the supplementary material.

\textbf{Evaluation.}
 We directly utilize the unsupervised pre-trained model as the feature extractor without any fine-tuning. Given the input frame with  spatial resolution of $H\times W$, the evaluation is realized on the res$_4$ layer with a spatial resolution of $\frac{H}{8} \times \frac{W}{8}$~(stride=8) or $\frac{H}{4} \times \frac{W}{4}$~(stride=4). To propagate the semantic labels from the initial ground-truth annotation, the recurrent inference strategy is applied following recent studies~\cite{jabri2020space,lai2020mast,xu2021rethinking}. More specifically,  the semantical label of the first frame, as well as previous predictions, are propagated to the current frame with the help of affinity computed between the feature maps of video frames. We evaluate our method over three downstream tasks including semi-supervised video object segmentation on DAVIS-2017~\cite{pont20172017}, human part propagation on VIP~\cite{zhou2018adaptive}, and pose keypoint tracking on JHMDB~\cite{jhuang2013towards}.
 
 \subsection{Ablation Studies}
 The ablation studies are performed with semi-supervised video object segmentation on the DAVIS-2017 validation set~\cite{pont20172017}. We use the mean of region similarity $\mathcal{J}_m$, mean of contour accuracy $\mathcal{F}_m$ and their average $\mathcal{J} \& \mathcal{F}_m$ as the evaluation metrics. We set the stride until layer res$_4$ to 8 for training and evaluation if not specified.

%  \textbf{The training data for spatial feature learning.} We first study how the training data affect spatial feature learning. Apart from training on ImageNet~\cite{deng2009imagenet}, we also train the model with $\mathcal{L}_{nce}$ on the video dataset (randomly sampling one frame in the video as a query). The results are shown in the first two rows of Table~\ref{tab:configurations}. Due to the existence of redundancy in space, the video dataset~\cite{xu2018youtube} tends to have less spatial diversity than the image dataset~\cite{deng2009imagenet}, which results in the unsatisfactory results~(60.1\% vs 66.5\%). This motivates us to switch to the image dataset for learning better appearance features.

 \textbf{Temporal feature learning.} We examine how each design in our temporal feature learning impacts the overall performance, which is shown in Table \ref{tab:abalations_temporal}. From the table, we can see leveraging more temporal cues with per-level frame reconstruction leads to an improvement in the range of 0.7\%. With the guidance of a better local correlation map at fine-grained pyramid level, $\mathcal{L}_{\mathrm{lc}}$ boosts up the accuracy from 65.6\% to 68.4\%. Moreover, enforcing the local correlation distillation to focus on the region with higher entropy leads to a performance gain in the range of 0.9\%. By fusing the above components, the performance reaches 69.3\%.

 \textbf{Different training configurations.} We then study the effect of three different ways as defined in Sec.~\ref{configurations}, which is shown in Table~\ref{tab:configurations}. 
 
 (i) Spatiotemporal. We use data from both ImageNet and YouTube-VOS for training and find the Spatiotemporal will result in unsatisfactory performance. This is attributed to the conflicts of two object functions in terms of training hyper-parameters and augmentations. More specifically, the contrastive model benefits from complicated data augmentations~\cite{chen2020improved} for learning invariant features within appearance changes. However, the pixel-level distortion may hinder the lower-level convolution features from finding the fine-grained motion between video frames~\cite{xu2021rethinking}.

 (ii) Spatial-then-Temporal. The above problem can be alleviated by our proposed Spatial-then-Temporal learning where each loss is optimized individually with its own hyper-parameters and augmentations. Our method without using $\mathcal{L}_{\mathrm{gc}}$ can already reach 69.4\%, and the $\mathcal{L}_{\mathrm{gc}}$ further boosts up the performance from 69.4\% to 71.3\%. Here we also compare our method with the general continual model EWC~\cite{kirkpatrick2017overcoming} based on weight constraints.  Our method still gets better results, which demonstrates the effectiveness of our proposed global correlation distillation.

 (iii) Temporal-then-Spatial. Directly fine-tuning the temporal model with $\mathcal{L}_{nce}$ decreases the performance from 67.5\% to 66.9\%. Unexpectedly, using EWC as a regularization strategy does not provide much benefit.

\begin{table}
\centering
\small
\resizebox{0.45\textwidth}{!}{
  \setlength\tabcolsep{3pt}
    \begin{tabular}{cccc|c}
    \thickhline
      %  \rowcolor{gray}
    Frame&Pyramid & \multicolumn{1}{c}{Local Corr.}& \multicolumn{1}{c|}{Entropy} & DAVIS  \\ 
      % \rowcolor{gray}
      \multirow{1}{*}{Recons.} & \multirow{1}{*}{Frame Recons.} &  \multirow{1}{*}{Distillation} & \multirow{1}{*}{Selection} & $\mathcal{J}$\&$\mathcal{F}_m$ $\uparrow$    \\ \hline
      $\checkmark$ &  && & 64.9 \\
      $\checkmark$ & $\checkmark$ & &  & 65.6 \\
      $\checkmark$ & $\checkmark$ & $\checkmark$ &  & 68.4  \\
      $\checkmark$ & $\checkmark$ & $\checkmark$ & $\checkmark$ &\textbf{69.3}  \\
    \thickhline
  \end{tabular}
}
\captionsetup{font=small}
\caption{\textbf{Ablation study for temporal feature learning with ResNet-18}. All models are learnt on YouTube-VOS~\cite{xu2018youtube} and evaluated on DAVIS-2017~\cite{pont20172017}.}
% \vspace{-3mm}
\label{tab:abalations_temporal}
\end{table}

\begin{table}[t]
  \centering
    \small
    \resizebox{0.47\textwidth}{!}{
		  \setlength\tabcolsep{9.3pt}
      %   \tablestyle{1pt}{1.1}
      \begin{tabular}{cccc}
        \thickhline
        \multirow{2}{*}{Configuration} & \multirow{2}{*}{Dataset} & \multirow{2}{*}{Loss Function} & \multirow{1}{*}{DAVIS} \\ 
        & & & $\mathcal{J} \& \mathcal{F}_m \uparrow$ \\
        \hline
        \multirow{1}{*}{Spatial only}  
        % & YTV                & $\mathcal{L}_{\mathrm{nce}}$                             & 60.1  \\
         & I                & $\mathcal{L}_{\mathrm{nce}}$                             & 66.5   \\ \hline
        \multirow{1}{*}{Temporal only} & YTV              & $\mathcal{L}_{\mathrm{t}}$                              & 67.5   \\ \hline
        \multirow{1}{*}{Spatiotemporal} 
        & I~+~YTV              & $\mathcal{L}_{\mathrm{nce}}$ + $\mathcal{L}_{\mathrm{t}}$                              & 68.6  \\ 
        \hline
        \multirow{3}{*}{\begin{tabular}[c]{@{}c@{}}Spatial-then-\\ Temporal\end{tabular}}
                                       & \multirow{3}{*}{I$\rightarrow$YTV}            & $\mathcal{L}_{\mathrm{nce}}$ $\rightarrow$ $\mathcal{L}_{\mathrm{t}}$          & 69.4   \\
                                      %  & I$\rightarrow$YTV            & $\mathcal{L}_{\mathrm{nce}}$ $\rightarrow$ $\mathcal{L}_{\mathrm{t}}$ + EWC    &  69.2  \\
                                       &          & $\mathcal{L}_{\mathrm{nce}}$ $\rightarrow$ $\mathcal{L}_{\mathrm{t}}$ + EWC    &  69.6  \\
                                       &             & $\mathcal{L}_{\mathrm{nce}}$ $\rightarrow$ $\mathcal{L}_{\mathrm{t}}$ + $\mathcal{L}_{\mathrm{gc}}$  & \textbf{71.3}   \\
      \hline
      \multirow{2}{*}{\begin{tabular}[c]{@{}c@{}}Temporal-then-\\ Spatial\end{tabular}} 
      & \multirow{2}{*}{YTV$\rightarrow$I}           &   $\mathcal{L}_{\mathrm{t}}$  $\rightarrow$ $\mathcal{L}_{\mathrm{nce}}$  & 66.9   \\
      &            &   $\mathcal{L}_{\mathrm{t}}$  $\rightarrow$ $\mathcal{L}_{\mathrm{nce}}$ + EWC  & 67.0   \\
       \thickhline
        \end{tabular}%
      }
      \captionsetup{font=small}
      \caption{\textbf{Ablation study for different training configurations with ResNet-50}. ``$\rightarrow$" indicates the switch in our two-step training in terms of datasets and loss functions. I: ImageNet~\cite{deng2009imagenet}.  YTV: YouTube-VOS~\cite{xu2018youtube}.}
      \label{tab:configurations}\vspace{-5mm}
\end{table}

% \begin{figure}[!tb]
%   \centering
%   {\includegraphics[width=0.48\textwidth]{figure/abalations/ablations2.pdf}}
%   \vspace{-6mm}
%   \caption{\small \textbf{The matching results for the query point.} The green/red bounding boxes indicate the correct/wrong matching.}
%   \label{fig:ablations}
%   \vspace{-6mm}
% \end{figure}

\begin{table*}[t]
  % \vspace{-0.5em}
	\centering
	\small
	\resizebox{0.83\textwidth}{!}{
		\setlength\tabcolsep{6.7pt}
		\renewcommand\arraystretch{1.00}
		\begin{tabular}{ccccccccc}
			\toprule
      & &  & & \multicolumn{2}{c}{Training Dataset} & & &  \\
      \cline{5-6}
			\multirow{-2}{*}{Method} & \multirow{-2}{*}{Sup.} & \multirow{-2}{*}{Backbone} & \multirow{-2}{*}{Stride} & Image & Video & \multirow{-2}{*}{$\mathcal{J}$\&$\mathcal{F}_m$ $\uparrow$} & \multirow{-2}{*}{$\mathcal{J}_m$ $\uparrow$}  &  \multirow{-2}{*}{$\mathcal{F}_m$ $\uparrow$}   \\  \hline
      MoCo~\cite{he2020momentum} & & ResNet-18 & 8 & ImageNet &-
			& 60.8 & 58.6  & 63.1  \\
      % DetCo & ResNet-18 & 8 & I &-
			% & 61.7 & 59.6  & 62.8  \\
      SimSiam~\cite{chen2021exploring} & & ResNet-18 & 8 & ImageNet &-
			& 62.0 & 60.0  & 64.0  \\
			Colorization~\cite{vondrick2018tracking} & & ResNet-18 & 8 & - & Kinetics
			& 34.0 & 34.6  & 32.7 \\
			CorrFlow~\cite{lai2019self} & & ResNet-18 & 8 & - & OxUvA
			& 50.3 & 48.4  & 52.2  \\
			MuG~\cite{lu2020learning} & & ResNet-18 &  8 & - & OxUvA
			& 54.3 & 52.6 & 56.1   \\
      UVC~\cite{li2019joint}  & & ResNet-18 & 8 & COCO & Kinetics
			& 59.5 & 57.7  & 61.3  \\
			ContrastCorr~\cite{wang2020contrastive} & & ResNet-18 & 8 & COCO  & TrackingNet
			& 63.0 & 60.5  & 65.5  \\
			VFS~\cite{xu2021rethinking}  & & ResNet-18 & 8 & - & Kinetics
			& 66.7 & 64.0  & 69.4  \\
      CRW~\cite{jabri2020space} & & ResNet-18 & 8 & - & Kinetics
			& 67.6 & 64.8  & 70.2  \\
			JSTG~\cite{zhao2021modelling} & & ResNet-18 & 8 & - & Kinetics
			& 68.7 & 65.8   & 71.6  \\
      CLSC~\cite{son2022contrastive} & & ResNet-18 & 8 & - & Kinetics
			& \underline{70.5} & \underline{67.4}   & \textbf{73.6}  \\
      SFC~\cite{hu2022semantic} & & ResNet-18 & 8 & - & YTV
			& 67.7 & 64.7   & 70.5  \\
      DUL~\cite{araslanov2021dense} & & ResNet-18 & 8 & - & YTV
			& 69.3 & 67.1   & 71.6  \\
      CLTC~\cite{jeon2021mining} & & ResNet-18 & 8 & - & YTV
			& 70.3 & 67.9   & 72.6  \\
			\textbf{Ours~(Temporal)} & & ResNet-18 & 8 & - & YTV
			& 69.3 & 66.7 & 71.9 \\
      \textbf{Ours} & & ResNet-18 & 8 & ImageNet & YTV
			& \textbf{70.7} & \textbf{68.0} & \underline{73.5} \\
      \hline
      MAST~\cite{lai2020mast} & & ResNet-18 & 4 & - & YTV
			& 65.5 & 63.3  & 67.6  \\
      MAMP~\cite{miao2022self} & & ResNet-18 & 4 & - & YTV
			& 69.7 & 68.3   & 71.2  \\
      LIIR~\cite{li2022locality} & & ResNet-18 & 4 & - & YTV
			& 72.1 & 69.7   & 74.5  \\
      \textbf{Ours~(Temporal)} & & ResNet-18 & 4 & - & YTV
			& \underline{72.5}  & \underline{69.9} & \underline{75.1} \\
      \textbf{Ours} & & ResNet-18 & 4 & ImageNet & YTV
			& \textbf{73.6} & \textbf{70.7} & \textbf{76.4} \\
			\hline
      MoCo~\cite{he2020momentum} & & ResNet-50 & 8 & ImageNet &-
			& 65.4 & 63.2  & 67.6  \\
      SimSiam~\cite{chen2021exploring} & & ResNet-50 & 8 & ImageNet &-
			& 66.3 & 64.5  & 68.2  \\
      % DetCo & ResNet-50 & 8 & I &-
			% & 66.5 & 64.7  & 68.4  \\
      TimeCycle~\cite{wang2019learning} & & ResNet-50 & 8 & - & VLOG
			& 40.7 & 41.9  & 39.4  \\
      UVC~\cite{li2019joint}  & & ResNet-50 & 8 & COCO & Kinetics
			& 56.3 & 54.5  & 58.1  \\
      % SeCo~\cite{yao2021seco} & & ResNet-50 & 8 & - & Kinetics
			% & 60.6 & 58.4  & 62.8  \\
      VINCE~\cite{gordon2020watching} & & ResNet-50 & 8 & - & Kinetics
			& 65.6 & 63.4  & 67.8  \\
      VFS~\cite{xu2021rethinking}  & & ResNet-50 & 8 & - & Kinetics
			& \underline{68.9} & \underline{66.5} & \underline{71.3}    \\
      \textbf{Ours} & & ResNet-50 & 8 & ImageNet & YTV
			& \textbf{71.3}  & \textbf{68.5} & \textbf{74.0}  \\
      % \hline
      % VFS~\cite{xu2021rethinking}  & & ResNet-50 & 4 & - & Kinetics
			% & \underline{68.9} & \underline{66.5} & \underline{71.3}    \\
      \textbf{Ours} & & ResNet-50 & 4 & ImageNet & YTV
			& \textbf{74.1}  & \textbf{71.1} & \textbf{77.1}  \\
      \hline
      SFC$^\dagger$~\cite{hu2022semantic} & & ResNet-18 + ResNet-50 & 8 & ImageNet & YTV
			& \underline{71.2} & \underline{68.3}   & \underline{74.0}  \\
      \textbf{Ours}$^\dagger$ & & ResNet-18 + ResNet-50 & 8 & ImageNet & YTV
			& \textbf{72.1}  & \textbf{69.5} & \textbf{74.7}  \\
      \hline
      Supervised~\cite{he2016deep} &$\checkmark$ & ResNet-18 & 8 & ImageNet & -
			& 62.9 & 60.6  & 65.2  \\
      Supervised~\cite{he2016deep} & $\checkmark$& ResNet-50 & 8 & ImageNet &-
			& 66.0 & 63.7  & 68.4  \\
      OnAVOS~\cite{voigtlaender2017online} & $\checkmark$ & ResNet-38 & - & I + C + P & D
			& 65.4 & 61.6  & 69.1  \\
      OSVOS-S~\cite{maninis2018video}  & $\checkmark$ & VGG-16 & - & I + P & D
			& 68.0   & 64.7 & 71.3  \\
      FEELVOS~\cite{voigtlaender2019feelvos}  & $\checkmark$ & Xception-65 & - & I + C & D + YTV
			& 71.5   & 69.1 & 74.0  \\
			\bottomrule
		\end{tabular}
	}
	\captionsetup{font=footnotesize}
	\caption{\textbf{Quantitative results for video object segmentation on validation set of DAVIS-2017}~\cite{pont20172017}. We show the results of state-of-the-art self-supervised methods and some supervised methods for comparison. Abbreviations for some datasets and their sizes -- number of images or duration of videos -- are: (h stands for hours). ~I:ImageNet~\cite{deng2009imagenet}~(1.28m). ~C:COCO~\cite{lin2014microsoft}~(30k). ~O:OxUvA~\cite{valmadre2018long}~(14h). ~T:TrackingNet~\cite{muller2018trackingnet}~(300h). ~K:Kinetics~\cite{carreira2017quo}~(800h). ~V:VLOG~\cite{fouhey2018lifestyle}~(344h). ~YTV:YouTube-VOS~\cite{xu2018youtube}~(5h). ~D:DAVIS-2017~\cite{pont20172017}. ~P:PASCAL-VOC~\cite{everingham2015pascal}.  $\dagger$ indicates using the late fusion~\cite{hu2022semantic} of two separate models for inference.}
	\label{table:sota}
	\vspace{-3mm}
\end{table*}

\textbf{Further analysis.} 
We give a further analysis here based on the above experiments. On the one hand, temporal feature learning helps to recognize the temporal pattern between video frames, which is unable to accomplish by training an appearance model. As can be seen in the last two rows of Figure \ref{fig:teaser}, the appearance model trained with $\mathcal{L}_{\mathrm{nce}}$ is misled by two patches at different locations which have a similar appearance, while the model trained with $\mathcal{L}_{\mathrm{t}}$ tends to learn better temporal representations~(see Temporal and Spatial-then-Temporal).  However, in the first two rows of Figure \ref{fig:teaser}, the model trained only with $\mathcal{L}_{\mathrm{t}}$ fails to get accurate correspondence when facing severe temporal discontinuity, e.g., occlusions, appearance changes, deformations,  while the model trained with $\mathcal{L}_{\mathrm{nce}}$ is able to correct the mistakes by tracking the points based on learned discriminative and robust spatial cues~(see Spatial and Spatial-then-Temporal).

\vspace{1mm}
\subsection{Comparison with State-of-the-Art Methods}
% \vspace{1mm}

\begin{figure*}[!t]
  \centering
  {\includegraphics[width=1.0\textwidth]{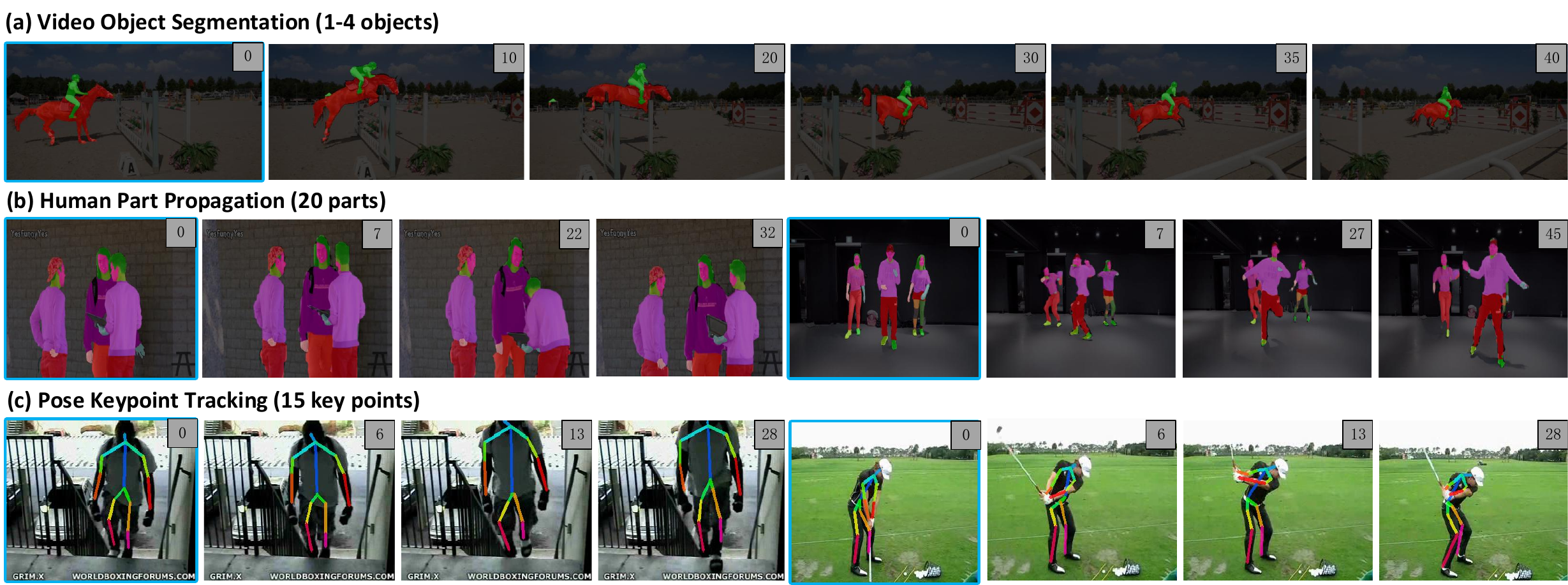}}
  \vspace{-3mm}
  \caption{\small Qualitative results for label propagation. Given the first frame~(blue border) with different annotations, we propagate it to the current frame without fine-tuning our correspondence model. (a) Video object segmentation on DAVIS-2017~\cite{pont20172017}. (b) Human part propagation on VIP~\cite{zhou2018adaptive}. (c) Pose keypoint tracking on JHMDB~\cite{jhuang2013towards}. }
  \vspace{-4mm}

  \label{fig:quan}

\end{figure*}

\textbf{Results for video object segmentation.}
We compare our method against previous self-supervised methods in Table~\ref{table:sota}. For a fair comparison, we report both results of setting the stride until layer res$_4$ to 4 and 8. Our method achieves state-of-the-art performance using both ResNet-18 and ResNet-50. For ResNet-18, our method with a stride of 8 achieves 70.7\%, surpassing all baselines using the same architecture. 
Benefiting from exploiting more fine-grained temporal cues for temporal feature learning by setting the stride to 4, the performance of our method reaches 73.6\%, leading to a performance gain of 1.5\% over LIIR~\cite{li2022locality}. For ResNet-50, our method still outperforms VFS~\cite{xu2021rethinking} by 2.4\%. It is worth noting that the studies in~\cite{jabri2020space, li2019joint, wang2020contrastive, xu2021rethinking, zhao2021modelling,son2022contrastive} are all pre-trained on large-scale video datasets, i.e., Kinetics~\cite{carreira2017quo}, TrackingNet~\cite{muller2018trackingnet}, while our method adopts a small video dataset plus an image dataset which has a smaller data size than video.  Besides, our model trained with temporal feature learning reaches 69.3\%/72.5\% and still leads the performance over other methods with the same training data~(YTV). Compared with SFC~\cite{hu2022semantic} which fuses learned semantic and fine-grained features with separate models, our single model with ResNet-50 already leads the performance, which demonstrates the importance of achieving synergy between spatial and temporal cues. Following the inference strategy of SFC, we fuse the spatial features~(ResNet-50)~learned with $\mathcal{L}_{nce}$ and temporal features~(ResNet-18)~learned with $\mathcal{L}_t$, the performance reaches 72.1\% and surpasses SFC by 0.9\%. More remarkably, Our method even outperforms some task-specific fully-supervised algorithms~\cite{maninis2018video, voigtlaender2017online, voigtlaender2019feelvos}.

\begin{table}
	\centering
	\small
	\resizebox{0.47\textwidth}{!}{
			\setlength\tabcolsep{2pt}
		\renewcommand\arraystretch{1.05}
		\begin{tabular}{ccccc}
            \toprule
		     & & \multicolumn{1}{c}{VIP} & \multicolumn{2}{c}{JHMDB} \\ \cline{3-5}
            \multirow{-2}{*}{Methods} & \multirow{-2}{*}{Sup.} & mIoU $\uparrow$   & PCK@0.1 $\uparrow$ & PCK@0.2 $\uparrow$
			\\ \midrule
      % ResNet-50 & $\checkmark$ & 39.5 & 59.2 & 78.3 \\
			TimeCycle~\cite{wang2019learning} &  & 28.9 & 57.3 & 78.1 \\
			UVC~\cite{li2019joint} &  & 34.1  & 58.6 & 79.6\\
      SFC~\cite{hu2022semantic} &  & 34.0  & 59.3 & 80.8\\
			CRW~\cite{jabri2020space} &  & 38.6  & 59.3 & 80.3 \\
			ContrastCorr~\cite{wang2020contrastive} &  & 37.4  & 61.1 & 80.8 \\
			VFS~\cite{xu2021rethinking}  &  & 39.9  & 60.5 & 79.5\\
			CLTC~\cite{jeon2021mining} &  & 37.8  & 60.5 & 82.3\\
			JSTG~\cite{zhao2021modelling} &  & 40.2  & 61.4 & \textbf{85.3}\\
      CLSC~\cite{son2022contrastive} &  & \underline{40.8}  & \underline{61.7} & 82.6\\
			\textbf{Ours} &  & \textbf{41.0} & \textbf{63.1} & \underline{82.9}\\
      \hline
      ResNet-18~\cite{he2016deep} & $\checkmark$ & 31.9 & 53.8 & 74.6 \\
      ATEN~\cite{zhou2018adaptive} & $\checkmark$ & 37.9 & - & -\\
      Thin-Slicing Net~\cite{song2017thin} & $\checkmark$ & - & 68.7 & 92.1\\
			\bottomrule
		\end{tabular}
	}
	\captionsetup{font=small}
	\caption{\textbf{Quantitative results for human part propagation and pose keypoint tracking.} We show results of state-of-the-art self-supervised methods and some supervised methods for comparison.}
	\label{table:vip}
	\vspace{-4mm}
\end{table}

\textbf{Results for human part propagation.} Next, we evaluate our method for human part tracking. Experiments are conducted on the validation set of VIP~\cite{zhou2018adaptive}, which consists of 50 videos with 19 human semantic parts, requiring more precise matching than DAVIS-2017~\cite{pont20172017}. Following~\cite{zhou2018adaptive}, we adopt mean intersection-over-union (mIoU) as the evaluation metric and resize the video to 560 $\times$ 560. All models are set to ResNet-18 with a stride of 8 for fair comparisons. The results are shown in Table \ref{table:vip}. Our method still surpasses previous state-of-the-art. Notably, our model outperforms ATEN~\cite{zhou2018adaptive} which is specifically designed for the task using human annotations. Figure \ref{fig:quan} (b) depicts some visualization results on several representative videos. Our approach could output tight boundaries around the multiple targets.

\textbf{Results for pose keypoint tracking.} We then make a performance comparison on the downstream task of human pose tracking. We conduct the experiments on the validation of JHMDB~\cite{jhuang2013towards} which has 268 videos. The annotations consist of 15 body joints for each person. The probability of correct keypoint~\cite{yang2012articulated} is utilized here to examine the accuracy with different thresholds. Following the evaluation protocol of~\cite{jabri2020space, li2019joint}, we resize the video frames to 320 $\times$ 320. The results in Table \ref{table:vip} show a consistent performance gain over previous methods, which successfully demonstrates the transferability of our method to different downstream tasks. The visualization results in Figure \ref{fig:quan}~(c) show the robustness of our approach to various challenges.

\vspace{1mm}

\section{Conclusions}
In this paper,  we present a new spatial-then-temporal pretext task for learning the representations of video correspondence.  The key idea underpinning the proposed method is to achieve the synergy between spatial and temporal cues. Specifically, we firstly train the model using contrastive loss with still images and then perform reconstructive learning to exploit temporal cues with a video dataset.  In the second step, we devise a global correlation distillation loss to retain the spatially-discriminative features learned in the first step. At the same time, we propose a local correlation distillation loss to alleviate the problem of temporal discontinuity. Extensive experiments on three kinds of downstream tasks validate the effectiveness of the proposed two-step task and loss functions. We hope that this work may provide a new perspective for video correspondence and brings us one step closer to the accurate correspondence for real-world videos.

%-------------------------------------------------------------------------
% \noindent\textbf{Limitations}. While previous studies suggest using large-scale video datasets for training, we only explore a small video dataset for realizing the temporal feature learning, considering the time-consuming training on Kinetics~\cite{carreira2017quo}. It is worthy of future work to figure out what the performance will be when scaling up to larger video datasets. 

%-------------------------------------------------------------------------
% For video correspondence, using high-resolution features for inference brings about unaffordable computational costs. However, we find temporal feature learning indeed benefits a lot from it due to more fine-grained temporal cues for reconstructive learning. Although local correlation distillation improves relatively coarse features and gets state-of-the-art results with stride of 4/8, there is still room for improvement.  Besides, 

%%%%%%%%% REFERENCES
{\small
\bibliographystyle{ieee_fullname}
\bibliography{egbib}
}

\end{document}